\documentclass[twocolumn]{cinc}
\usepackage[utf8]{inputenc}                     
\usepackage{blindtext}
\usepackage[T1]{fontenc}
\usepackage{graphicx}                           
\usepackage{rotating}                           
\usepackage{pdflscape}                          
\usepackage{verbatim}                           
\usepackage{url}                                
\usepackage{indentfirst}                        
\usepackage{float}                              
\usepackage{breakcites}
\usepackage{amsmath}                            
\usepackage{changepage}                         
\usepackage{cancel}
\usepackage{listings}
\usepackage{wrapfig}
\usepackage{enumitem}                           
\usepackage{array}                              
\usepackage{makecell}                           
\usepackage{amssymb}                            
\usepackage[table]{xcolor}
\usepackage{mathtools}
\usepackage{soul}
\usepackage{graphicx}
\usepackage{amsmath, amssymb}
\usepackage{hyperref}
\usepackage{algorithm}
\usepackage{booktabs}
\usepackage{subcaption}
\usepackage{multirow}
\usepackage{tikz} 
\usepackage{cite}
\usepackage{setspace}

\begin{document}
\bibliographystyle{cinc}

\title{SAF-Net: Self-Attention Fusion Network for \\ Myocardial Infarction Detection using Multi-View Echocardiography}


\author {Ilke Adalioglu$^{1}$, Mete Ahishali$^{1}$, Aysen Degerli$^{1}$, Serkan Kiranyaz$^{2}$, Moncef Gabbouj$^{1}$ \\
\ \\ 
 $^1$ Tampere University, Tampere, Finland \\
$^2$  Qatar University, Doha, Qatar  }
\maketitle

\begin{abstract}
\vspace{-0.15cm}
%
%
    Myocardial infarction (MI) is a severe case of coronary artery disease (CAD) and ultimately, its detection is substantial to prevent progressive damage to the myocardium. In this study, we propose a novel view-fusion model named self-attention fusion network (SAF-Net) to detect MI from multi-view echocardiography recordings.
    The proposed framework utilizes apical 2-chamber (A2C) and apical 4-chamber (A4C) view echocardiography recordings for classification. Three reference frames are extracted from each recording of both views and deployed pre-trained deep networks to extract highly representative features.
    The SAF-Net model utilizes a self-attention mechanism to learn dependencies in extracted feature vectors. 
    The proposed model is computationally efficient thanks to its compact architecture having three main parts: a feature embedding to reduce dimensionality, self-attention for view-pooling, and dense layers for the classification.
    Experimental evaluation is performed using the HMC-QU-TAU\footnote{The benchmark HMC-QU-TAU dataset is publicly shared at the repository \href{https://www.kaggle.com/aysendegerli/hmcqu-dataset}{https://www.kaggle.com/aysendegerli/hmcqu-dataset.}} dataset which consists of 160 patients with A2C and A4C view echocardiography recordings. The proposed SAF-Net model achieves a high-performance level with 88.26\% precision, 77.64\% sensitivity, and 78.13\% accuracy. 
    The results demonstrate that SAF-Net model achieves the most accurate MI detection over multi-view echocardiography recordings.

\end{abstract}





\vspace{-0.25cm}
\section{Introduction}
\vspace{-0.1cm}




Early detection of Myocardial Infarction (MI), an indicator of coronary artery disease (CAD), is substantial to avoid further health complications. Echocardiography (echo) is a primary diagnostic tool used in MI investigation. Particularly, detection of the early signs of MI is often not viable using an electrocardiogram (ECG) as the disease has already progressed significantly by the time it becomes visible in ECG recordings \cite{taban2019value}.

Recently, there has been a growing interest in leveraging machine learning (ML) techniques in echo analysis to develop automated, fast, accessible, and cost-effective tools \cite{LowQ_echo, LvwallEstimation} to assist medical doctors (MDs). Nevertheless, there is a lack of research on MI detection and publicly accessible echo datasets are limited. In fact, only the study in \cite{LvwallEstimation} has released the first publicly available dataset containing apical 4-chamber (A4C) view echo recordings for the detection of MI. The proposed ML-based detection framework in \cite{LvwallEstimation} utilizes only single-view recordings. Obviously, if the infarcted cardiac muscle segment is invisible in the chamber view, it will be missed by the detection framework. To this end, the proposed approach in \cite{degerli2023early} focuses on capturing early MI from multi-view echo using displacement features extracted by analyzing segmented endocardial boundary.





As a recent trend, the attention mechanism introduced in \cite{vaswani2017attention} is considerably used in various domains achieving promising outcomes in challenging computer vision tasks such as image recognition, captioning, and recognition. By operating between different parts of the input or using learned context vectors, the attention mechanism is used to make predictions from different sequences \cite{Rao_2021_ICCV, Huang_2019_ICCV, zhao2020exploring}. Furthermore, the self-attention mechanism enables the model to capture long-term dependencies and contributions of the different features of different positions in a single sequence. The medical field benefits from self-attention with its capability of extracting robust and descriptive features by exploiting patterns within the data \cite{zhang2021transfuse, dong2023polyppvt3}. The ability to extract these descriptive features improved the performance in medical classification problems such as in \cite{wang2020attention, shi2020decoupled}. Similarly, self-attention is utilized to detect pulmonary nodule classification in \cite{Zhu_2022_ACCV} using multi-view data.

In this study, we propose a self-attention fusion network (SAF-Net) for early MI detection using multi-view echo recordings.  The proposed detection framework consists of a novel feature extraction technique that can capture the dependencies in the pruned feature space using a self-attention module. In summary, our contributions can be listed as follows: 
\begin{itemize}[leftmargin=0.4cm]
    \item We devise a framework for capturing salient features from multi-view echo recordings that reflect the important patterns. Our approach includes the extraction of three distinct frames in both views, complemented by the utilization of different state-of-the-art networks to extract the feature set.
    \item By utilizing the self-attention mechanism in the proposed SAF-Net, we improve the model's ability to learn the dependencies of extracted feature vectors, ultimately improving the detection of early MI.
    \item We conduct an extensive set of experimental evaluations for the comparison of the proposed approach against various classifiers. Accordingly, the proposed approach achieves the highest MI detection performance.
    \item We extend the multi-view echocardiography dataset, HMC-QU-TAU by incorporating data from 160 patients. 
\end{itemize}
\vspace{-0.2cm}
\section{Proposed Method}
\vspace{-0.1cm}

\begin{figure}[b!]
    \centering
    \includegraphics[width=7.9cm]{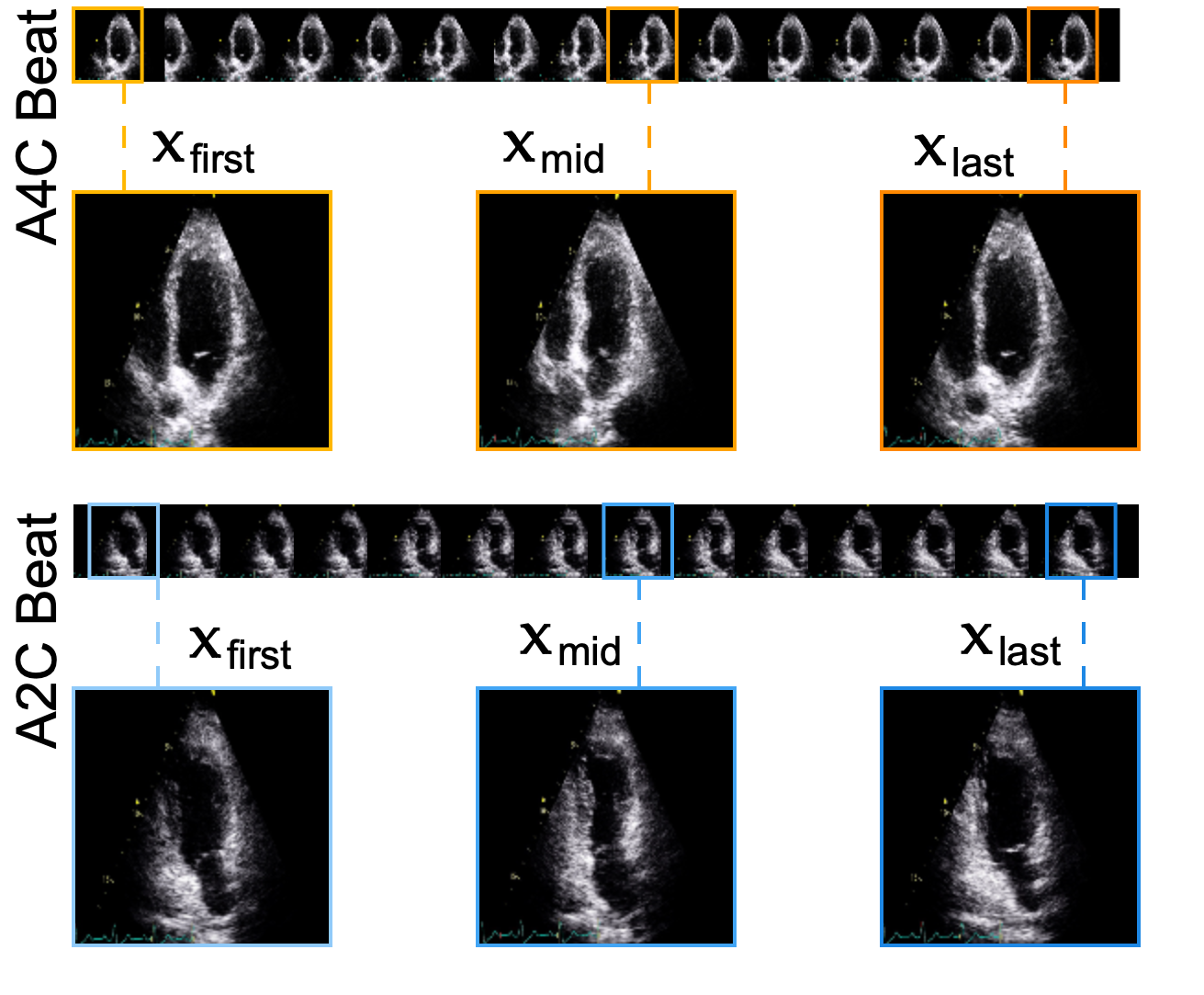}
    \caption{From one-cardiac cycle of both apical chamber views, three reference frames are extracted as the first, middle, and last.}
    \label{fig:beat-selection}
\end{figure}


The model performance is highly correlated to the descriptive power of the used features for the classification. In this manner, we propose the following feature extraction scheme. First, we extract the first, middle, and last frames in one-cardiac cycle from both A4C and A2C views for every recording. Let the selected reference frames be $\mathbf{X}_{\text{A4C}}$ and $\mathbf{X}_{\text{A2C}}$ for both views, where $\mathbf{X} \in $$\mathbb{R}$$^{N \times N\times 3}$ and $N = 224$. Then, for each recording, we utilize pre-trained networks initialized with their ImageNet \cite{5206848imagenet}  weights in order to extract representative features for each one-cardiac cycle. The selected frames are depicted in Figure \ref{fig:beat-selection}. Accordingly, we use the following state-of-the-art deep networks for the feature extraction given as $\mathbf{X}_{\text{A4C}}$ and $\mathbf{X}_{\text{A2C}}$ inputs:
\begin{itemize}[leftmargin=0.4cm]
    \item[-] \textbf{DenseNet-121} \cite{huang2018denselyDenseNet} is characterized by its fully-connected convolutional layers with skip-connections resulting in an effective information flow through the layers.
    \item[-] \textbf{Inception-v3} \cite{szegedy2015rethinkingInception} is a deep network that is designed for image classification and recognition. By utilizing parallel convolutions in modules, and keeping the dimensionality low, Inception can achieve promising performance levels with less computational complexity compared to the deeper competitors.
    \item[-] \textbf{ResNet-50} \cite{he2015deepResNet} mainly utilizes skip-connections to build a more stable but deeper network. It aims to overcome the problem of gradient vanishing by utilizing residual connections with addition operations.

\end{itemize}


\begin{figure}[b!]
    \centering
    \includegraphics[width=7.9cm]{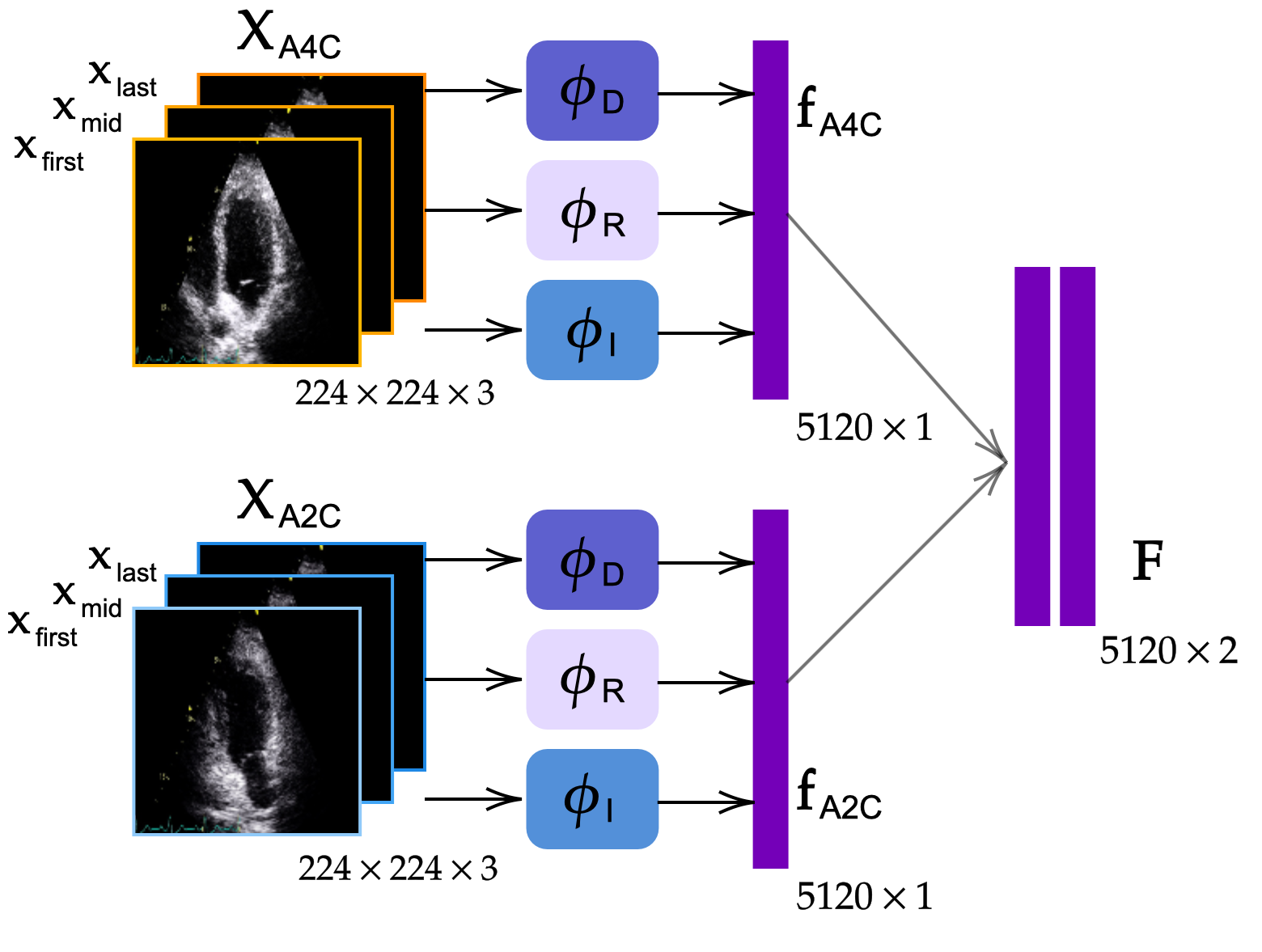}
    \caption{Features are extracted from both views frame set using pre-trained state-of-the-art networks: DenseNet-121 ($\phi_D$), ResNet-50 ($\phi_R$), and Inception-v3 ($\phi_I$).}
    \label{fig:feature-extraction}
\end{figure}

By using pre-trained ImageNet weights of the above-mentioned state-of-the-art deep networks, we aim to extract representative features to ensure state-of-the-art MI detection performance. Given a chamber view $\mathbf{X}$ for a patient, a pre-trained network provides the following mapping: $\phi: $$\mathbb{R}$$^{N \times N\times 3} \rightarrow  $$\mathbb{R}$$^{d}$ where $d$ is the feature vector dimension. The extracted features $\mathbf{f} = \phi(\mathbf{X})$ have a dimension of $d=1024$ for DenseNet-121, whereas $d=2048$ for Inception-v3 and ResNet-50. Then, as shown in Figure \ref{fig:feature-extraction}, we obtain the final concatenated feature vector as $\mathbf{f} = [f_D, f_I, f_R] \in $$\mathbb{R}$$^{5120}$. The feature pair for both chamber views containing $\mathbf{f}_{A2C}$ and $\mathbf{f}_{A4C}$ forms the matrix $F^{5120\times2}$ that is given to the SAF-Net model as input.

The first part of the SAF-Net consists of a feature embedding step for dimensionality reduction. For obtaining latent feature space $(L^{64\times2})$, we use one fully-connected layer with the Rectified Linear Unit (ReLU) activation function. After the feature extraction processes, we strategically placed the self-attention mechanism in the model. By using the mapped latent feature representation as input, the self-attention mechanism utilizes the benefits of capturing dependencies between the extracted and pruned features.

\begin{figure}[t!]
    \centering
    \includegraphics[width=7.9cm]{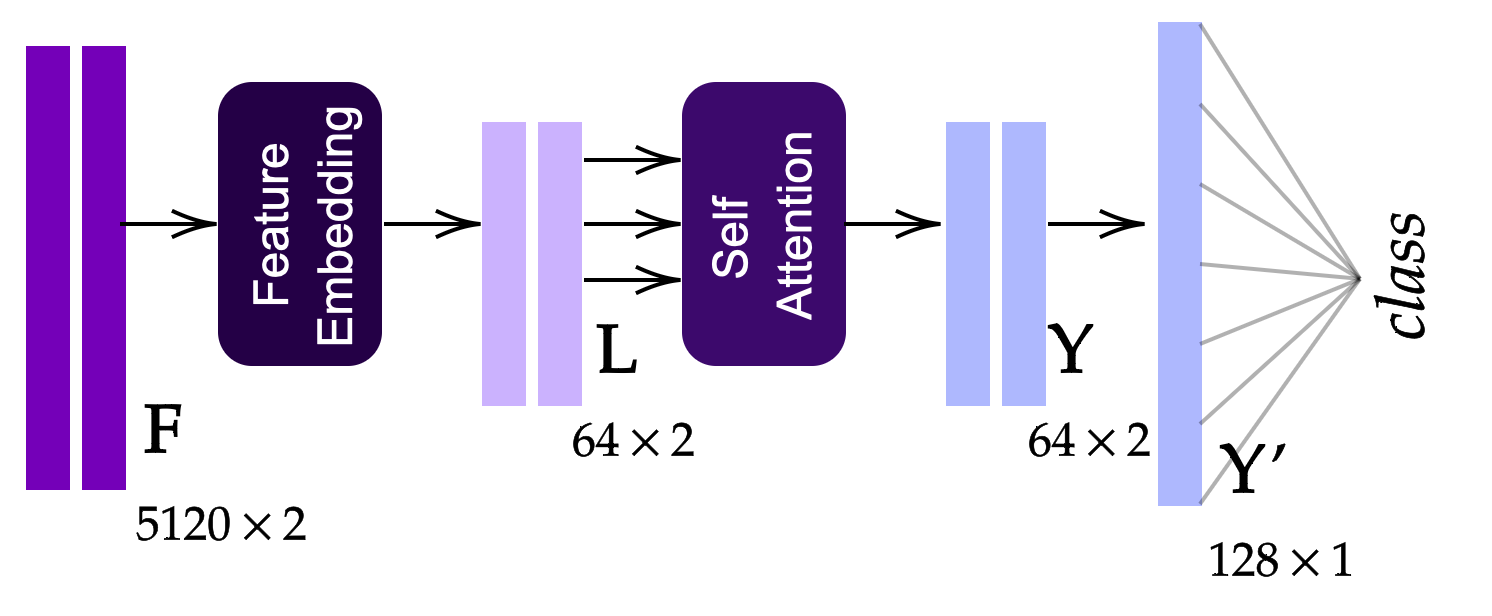}
    \caption{The proposed SAF-Net model consists of a feature embedding layer that computes latent space representation, $L$ of the feature matrix, $\mathbf{F}$ which is followed by a self-attention layer. Lastly, a fully-connected layer is attached to perform the classification.}
    \label{fig:classification}
\end{figure}

In the self-attention mechanism, for an input matrix $\mathbf{X} \in $$\mathbb{R}$$ ^{n \times d_{model}}$ where $n$ is the latent dimension after linear embedding layer and $d_{model}$ is the model's depth, the self-attention computation is performed using the weight matrices $\mathbf{W}_\textit{Q}$, $\mathbf{W}_\textit{K}$, and $\mathbf{W}_\textit{V}$ $\in  $$\mathbb{R}$$ ^{d_{model}\times d_k }$, where $d_k$ is the model dimension which is less than $d_k$. First, we calculate query $\mathbf{Q}$, key $\mathbf{K}$, and value $\mathbf{V}$ as $\mathbf{X}\mathbf{W}_\textit{Q} = \mathbf{Q}$, $\mathbf{X}\mathbf{W}_\textit{K} = \mathbf{K}$, and $\mathbf{X}\mathbf{W}_\textit{V} = \mathbf{V}$, where $\mathbf{Q},\mathbf{K}, \mathbf{V} \in $$\mathbb{R}$$^{n\times d_k} $. The weight matrices are learned through training to ensure a common-space representation for both views. By using the computed $Q$, $K$, and $V$, we calculate the attention scores as follows:
\vspace{-0.15cm}
\begin{equation}
    \text{Attention}(\mathbf{Q}, \mathbf{K}, \mathbf{V}) = \text{softmax}\left(\frac{ \mathbf{Q} \mathbf{K}^\textit{T}}{\sqrt{\textit{d}_\textit{k}}}\right) \mathbf{V},
    \label{attn}
\end{equation}
\vspace{-0.15cm}
where $d_k$ is used as a scaling factor. Essentially, Eq. \eqref{attn} is a scaled dot product that calculates the similarity between the vectors. Hence, we try to capture the relationship and dependencies between the embedded features. 
The last step in the MI classification process is to employ a fully-connected layer with a sigmoid activation function, subsequent to the self-attention layer. The proposed SAF-Net model is illustrated in Figure \ref{fig:classification}.


\vspace{-0.1cm}
\section{Experimental Evaluation}
\vspace{-0.1cm}
\textbf{Experimental Setup.} We use the HMC-QU-TAU dataset \cite{degerli2023early} to evaluate the performance of the proposed SAF-Net. The dataset consists of 320 echocardiography recordings of A2C and A4C views from 160 individuals; \textbf{103} MI patients and \textbf{57} of non-MI subjects. Considering the positive label as MI, the performance of the proposed and competing methods are computed using metrics as follows: Sensitivity (Sen) is the ratio of correctly predicted positives to all actual positives, Specificity (Spe) is the ratio of correctly predicted negatives to all actual negatives, Precision (Pre) is the ratio of correctly predicted positives to all predicted positives, F1-Score (F1) is the harmonic mean of Sen and Pre, Accuracy (Acc) is the correctly classified samples to all samples in dataset, and Geometric Mean (GM) is the geometric mean of Sen and Pre. We perform 10-fold cross-validation and report averaged performance. To train the classification and feature embedding layer, we use the Adam optimizer \cite{kingma2017adam} with a learning rate of $10^{-3}$ and weighted binary cross-entropy loss to overcome the challenges caused by the imbalanced dataset.

We have evaluated several methods for comparison including random forest (RF), decision tree (DT), k-nearest neighbor (KNN), support vector machine (SVM), and multi-layer perceptron (MLP) classifiers. We have performed grid-search approach for the hyper-parameter selection by performing another 3-fold CV over the training set. All classifiers are fed with the feature matrix $\mathbf{F}$ for a fair comparison with the proposed approach. The search parameters of SVM include linear, Gaussian, and polynomial kernels, where $C \in \{0.001, 0.01, 0.1, 1, 10, 100, 1000\}$, $\gamma \in \{0.001, 0.01, 0.1, 1, 10, 100, 1000\}$, and the degree of polynomial is set to $\{2, 3, 4\}$, respectively. The MLP classifier has the same structure as used in \cite{Ahishali_2021} and it is trained with Adam optimizer with a learning rate of $10^{-3}$ and binary cross entropy loss.

\vspace{-0.2 cm}
\begin{table}[H]
\caption{\label{tab:font} Averaged MI detection performance results ($\%$) obtained by the proposed SAF-Net approach and compared methods over the HMC-QU-TAU dataset.}
\resizebox{\columnwidth}{!}{
\begin{tabular}{@{}lllllll@{}}
\toprule
Model  & Sen            & Spe         & Pre            & F1             & Acc            & GM             \\ \midrule
SAF-Net & 77.64          & \textbf{79.0} & \textbf{88.26} & 81.57          & \textbf{78.13} & \textbf{77.19} \\
RF     & 88.36          & 52.33       & 78.37          & \textbf{82.11} & 75.62          & 66.49          \\
DT     & 66.18          & 51.33       & 72.52          & 67.28          & 60.63          & 56.52          \\
KNN    & 82.54          & 60.33       & 79.21          & 80.49          & 74.37          & 69.69          \\
SVM    & \textbf{90.09} & 39.66       & 75.30          & 80.82          & 72.50          & 43.90          \\ 
MLP & 76.63          & 64.66       & 81.45          & 77.47          & 72.50          & 67.73          \\\bottomrule
\end{tabular}
}
\label{tab:res}
\end{table}
\vspace{-0.2cm}
\textbf{Results and Discussion.} Table \ref{tab:res} illustrates the superiority of SAF-Net over widely recognized classifiers in terms of specificity, precision, accuracy, and geometric mean for MI detection. Despite its outstanding performance in various evaluation metrics, SAF-Net is unable to achieve the highest sensitivity score. On the other hand, the compared classifiers fail to achieve any reasonable specificity level. Especially, SAF-Net excels in specificity score, which shows its capacity to minimize false positives as it can be seen in Table \ref{tab:CMs}, it has achieved the highest GM value of $77.19\%$ with a sensitivity level of $77.64\%$. The performance gap in GM is significant by $7.5\%$ between the proposed approach and the closest competing method showing that the classifier has achieved the highest MI detection performance.  

\begin{table}[t!]
\centering
\caption{Cumulative confusion matrices of the proposed (a) SAF-Net, (b) RF, (c) DT, (d) KNN, (e) SVM, and (f) MLP models over test sets of each fold.}
\begin{subtable}{.48\linewidth}
\centering
\caption{}
\vspace{-0.2cm}
\resizebox{\linewidth}{!}{
\begin{tabular}{|c|c|c|c|}
\hline
\multicolumn{2}{|c|}{\multirow{2}{*}{\textbf{SAF-Net}}} & \multicolumn{2}{c|}{Predicted} \\ \cline{3-4} 
\multicolumn{2}{|c|}{} & \multicolumn{1}{c|}{Non-MI} & \multicolumn{1}{c|}{MI} \\ \hline
\multirow{2}{*}{\begin{tabular}[c]{@{}c@{}}Ground\\ Truth\end{tabular}} & Non-MI & 45 & 12 \\ \cline{2-4} 
 & MI & 23 & 80 \\ \hline
\end{tabular}}
\end{subtable}
\bigskip
\noindent
\begin{subtable}{.48\linewidth}
\centering
\caption{}
\vspace{-0.2cm}
\resizebox{\linewidth}{!}{
\begin{tabular}{|c|c|c|c|}
\hline
\multicolumn{2}{|c|}{\multirow{2}{*}{\textbf{RF}}} & \multicolumn{2}{c|}{Predicted} \\ \cline{3-4} 
\multicolumn{2}{|c|}{} & \multicolumn{1}{c|}{Non-MI} & \multicolumn{1}{c|}{MI} \\ \hline
\multirow{2}{*}{\begin{tabular}[c]{@{}c@{}}Ground\\ Truth\end{tabular}} & Non-MI & 30 & 27 \\ \cline{2-4} 
 & MI & 12 & 91 \\ \hline
\end{tabular}}
\end{subtable}

\begin{subtable}{.48\linewidth}
\centering
\caption{}
\vspace{-0.2cm}
\resizebox{\linewidth}{!}{
\begin{tabular}{|c|c|c|c|}
\hline
\multicolumn{2}{|c|}{\multirow{2}{*}{\textbf{DT}}} & \multicolumn{2}{c|}{Predicted} \\ \cline{3-4} 
\multicolumn{2}{|c|}{} & \multicolumn{1}{c|}{Non-MI} & \multicolumn{1}{c|}{MI} \\ \hline
\multirow{2}{*}{\begin{tabular}[c]{@{}c@{}}Ground\\ Truth\end{tabular}} & Non-MI & 29 & 28 \\ \cline{2-4} 
 & MI & 35 & 68 \\ \hline
\end{tabular}}
\end{subtable}
\bigskip
\noindent
\begin{subtable}{.48\linewidth}
\centering
\caption{}
\vspace{-0.2cm}
\resizebox{\linewidth}{!}{
\begin{tabular}{|c|c|c|c|}
\hline
\multicolumn{2}{|c|}{\multirow{2}{*}{\textbf{KNN}}} & \multicolumn{2}{c|}{Predicted} \\ \cline{3-4} 
\multicolumn{2}{|c|}{} & \multicolumn{1}{c|}{Non-MI} & \multicolumn{1}{c|}{MI} \\ \hline
\multirow{2}{*}{\begin{tabular}[c]{@{}c@{}}Ground\\ Truth\end{tabular}} & Non-MI & 34 & 23 \\ \cline{2-4} 
 & MI & 18 & 85 \\ \hline
\end{tabular}}
\end{subtable}

\begin{subtable}{.48\linewidth}
\centering
\caption{}
\vspace{-0.2cm}
\resizebox{\linewidth}{!}{
\begin{tabular}{|c|c|c|c|}
\hline
\multicolumn{2}{|c|}{\multirow{2}{*}{\textbf{SVM}}} & \multicolumn{2}{c|}{Predicted} \\ \cline{3-4} 
\multicolumn{2}{|c|}{} & \multicolumn{1}{c|}{Non-MI} & \multicolumn{1}{c|}{MI} \\ \hline
\multirow{2}{*}{\begin{tabular}[c]{@{}c@{}}Ground\\ Truth\end{tabular}} & Non-MI & 23 & 34 \\ \cline{2-4} 
 & MI & 10 & 93 \\ \hline
\end{tabular}}
\end{subtable}
\bigskip
\noindent
\begin{subtable}{.48\linewidth}
\centering
\caption{}
\vspace{-0.2cm}
\resizebox{\linewidth}{!}{
\begin{tabular}{|c|c|c|c|}
\hline
\multicolumn{2}{|c|}{\multirow{2}{*}{\textbf{MLP}}} & \multicolumn{2}{c|}{Predicted} \\ \cline{3-4} 
\multicolumn{2}{|c|}{} & \multicolumn{1}{c|}{Non-MI} & \multicolumn{1}{c|}{MI} \\ \hline
\multirow{2}{*}{\begin{tabular}[c]{@{}c@{}}Ground\\ Truth\end{tabular}} & Non-MI & 37 & 20 \\ \cline{2-4} 
 & MI & 24 & 79 \\ \hline
\end{tabular}}
\end{subtable}

\label{tab:CMs}
\end{table}
\vspace{-0.15cm}
\section{Conclusion}
\vspace{-0.1cm}
Early diagnosis of MI holds paramount significance to prevent further necrosis. Using a multi-view approach improves the identification accuracy, as it provides more visibility over myocardial segments for detecting MI. In this study, we proposed a novel feature extraction technique with frame selection for one-cardiac cycle by deploying pre-trained networks. Subsequently, the proposed SAF-Net model consisting of linear embedding, view pooling, and classification parts utilizing self-attention mechanisms exploits the dependencies in extracted features to maximize the MI detection performance. The experimental evaluations over the HMC-QU-TAU dataset consisting of multi-view echo pairs of 160 patients show that the proposed approach has achieved the highest accuracy, precision, specificity, and geometric mean with $78\%$, $88.26\%$, $79\%$, and $77.19\%$, respectively, as outperforming the competing well-known classifiers. This performance holds promise for the potential of SAF-Net in early MI detection and aiding healthcare professionals to enable their diagnostic tasks more effectively. Finally, the released HMC-QU-TAU dataset with this study lays the groundwork for future cardiology studies and our aspiration to leverage the potential of machine learning in the healthcare domain.


\balance

\vspace{-0.15cm}
\section*{Acknowledgments}  
\vspace{-0.15cm}
%
This work has been supported by Business Finland project AMALIA under NSF IUCRC Center for Big Learning and the Tietoevry Veturi program.


\begin{spacing}{0.85}
\vspace{-0.15cm}
\bibliography{refs}
\end{spacing}


  
  
      

\vspace{-0.5 cm}
\begin{correspondence}
Ilke Adalioglu\\ 
P.O. Box 553, FI-33014, Tampere Finland\\
ilke.adalioglu@tuni.fi
\end{correspondence}

\end{document}